\documentclass[10pt,twocolumn,letterpaper]{article}

\usepackage{cvpr}              
\usepackage{algorithm}
\usepackage{algorithmic}
\usepackage{amsmath} 
\usepackage{tabularray}
\usepackage[accsupp]{axessibility}
\usepackage{newfloat}
\usepackage{listings}
\definecolor{cvprblue}{rgb}{0.21,0.49,0.74}
\usepackage[pagebackref,breaklinks,colorlinks,allcolors=cvprblue]{hyperref}

\def\paperID{5434} 
\def\confName{CVPR}
\def\confYear{2026}

\title{PestVL-Net: Enabling Multimodal Pest Learning via Fine-grained Vision-Language Interaction}

\author{Xueheng Li$^{1,2}$\thanks{Equal contribution.}, 
Tao Hu$^{1,2}$\footnotemark[1], 
Ke Cao$^{1, 2}$, 
Runsheng Qi$^{1}$, 
Huixin Zhang$^{1, 2}$,\\
Rui Li$^{1}$, 
Jie Zhang$^{1, 3}$, 
Chengjun Xie$^{1, 3}$\thanks{Corresponding author.}\\
$^{1}$Institute of Intelligent Machines, Hefei Institutes of Physical Science, Chinese Academy of Sciences\\
$^{2}$University of Science and Technology of China\\
$^{3}$Zhongke Hefei Institute of Technology Innovation Engineering\\
{\tt\small \{lixueheng,ht\_simon\}@mail.ustc.edu.cn,}
{\tt\small \{lirui,zhangjie,cjxie\}@iim.ac.cn}
}

\begin{document}
\maketitle
\begin{abstract}
Effective pest recognition and management are crucial for sustainable agricultural development. However, collecting pest data in real scenarios is often challenging. Compared to other domains, pests exhibit a wide variety of species with complex and diverse morphological characteristics. Existing techniques struggle to effectively model the key visual and high-level semantic features of pests in a fine-grained manner. These limitations hinder the practical application of such methods in real agricultural scenarios.
To address these critical challenges, we  present a synergistic approach  that integrates PestVL-Net, a novel vision-language framework, with two multi-species pest datasets to facilitate fine-grained pest learning. The visual pathway of PestVL-Net utilizes the Recurrent Weighted Key Value (RWKV) architecture, incorporating a saliency-guided adaptive window partitioning scheme to effectively model the fine-grained visual characteristics of pests. Concurrently, the linguistic component generates precise pest semantic descriptions by leveraging Multimodal Large Language Models (MLLMs) priors, critically informed by agricultural expert knowledge and structured via multimodal Chain-of-Thought (CoT) reasoning. The deep fusion of these complementary visual and textual representations enables fine-grained multimodal pest learning. Extensive experimental evaluations on multiple pest datasets validate the superior performance of PestVL-Net, highlighting its potential for effective real-world pest management.
\end{abstract}

\section{Introduction}
\label{sec:intro}
The mounting challenge of pest-induced agricultural losses critically impacts global food security and sustainable farming practices \cite{anderson2019genetically}.  Effective mitigation strategies hinge critically upon the accurate identification of a diverse array of pest species, often manifesting as complex assemblages of several to dozens of distinct types within specific agricultural regions \cite{courson2022weather}. Therefore, the development of precise pest recognition and management frameworks plays a key role in agricultural productivity \cite{liu2021plant}. 


\begin{figure}[t]
	\centering
        \includegraphics[width=\linewidth]{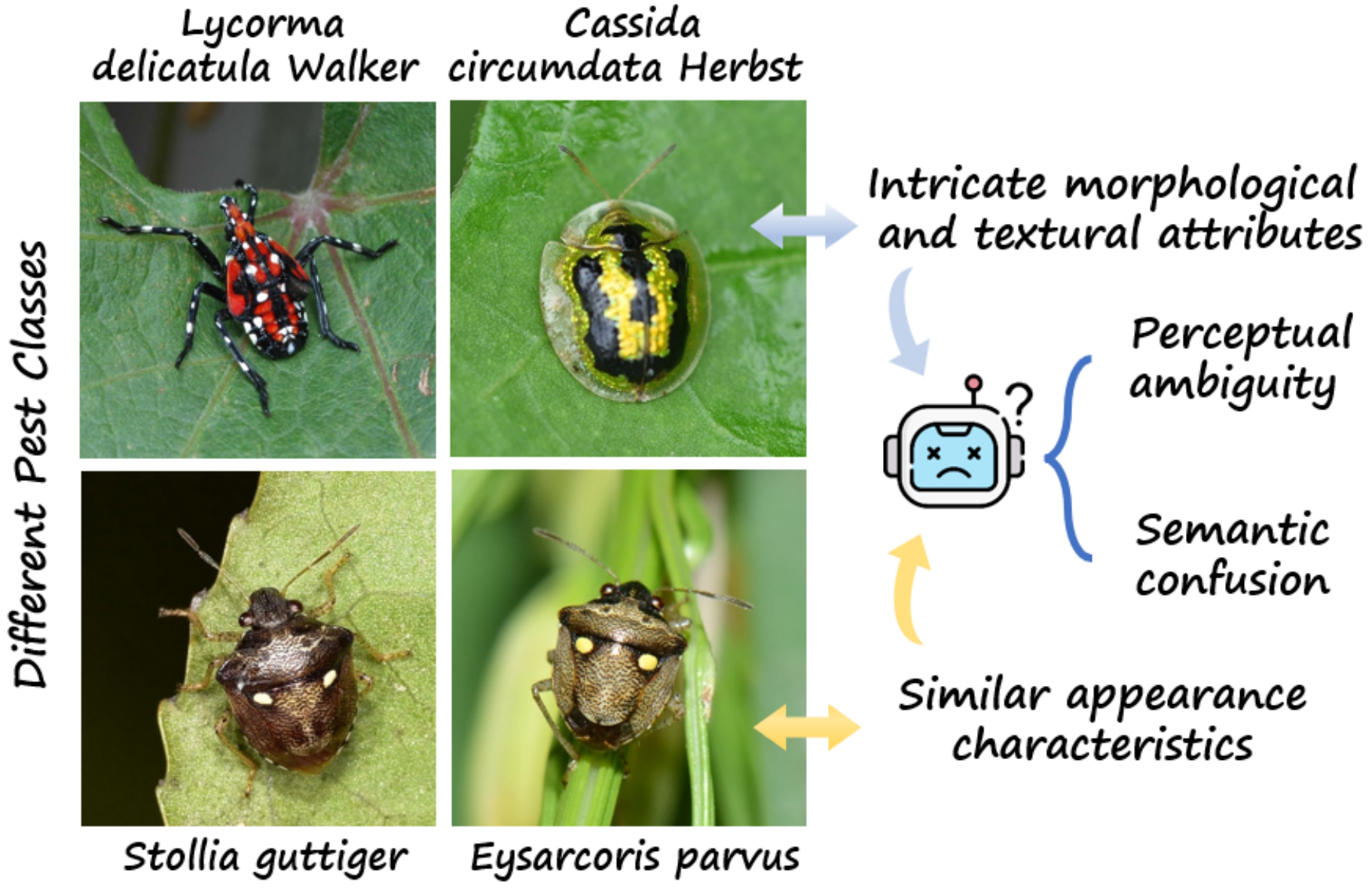}
	\caption{The intricate morphology and texture of pests, combined with high inter-class visual similarity, present significant challenges for effective model learning.}
\label{FIG:pest}
\end{figure}

In recent years, the rapid advancement of deep learning has significantly accelerated its adoption in smart agriculture applications \cite{liu2024auto}. While some recent efforts have focused on building large-scale insect datasets to support fundamental entomological research \cite{jin2025ip, shinoda2025agrobench}, such initiatives are often less aligned with the practical requirements of real-world agricultural systems.  In actual deployment scenarios, constrained computational resources on edge devices make it impractical to process large-scale data using resource-intensive models. Furthermore, agricultural applications typically demand accurate  recognition of a limited set of common crop pests, rather than exhaustive classification across the entire spectrum of insect species \cite{deguine2021integrated}.

Existing prevalent network architectures applied in the agricultural scene also exhibit notable limitations, such as local receptive fields  \cite{luo2016understanding} or high computational overhead \cite{kitaev2020reformer}, which constrain their efficiency in agricultural scenarios.
Recent research \cite{yuan2024mamba, yang2025restore} highlights the efficacy of the lightweight model, \textbf{R}ecurrent \textbf{W}eighted \textbf{K}ey \textbf{V}alue (RWKV), which has shown superior performance over Transformers in multiple domains \cite{duan2024vision}, owing to its efficient global receptive field and low-complexity attention mechanism. However, prevailing visual RWKV \cite{du2024exploring} primarily relies on horizontal or vertical scanning patterns and apply uniform processing to image patches. This strategy hinders the model’s ability to capture the essential features of pests, which are characterized by complex morphology and high textural heterogeneity (as shown in Fig. \ref{FIG:pest}).
\begin{figure}[t]
	\centering
        \includegraphics[width=\linewidth]{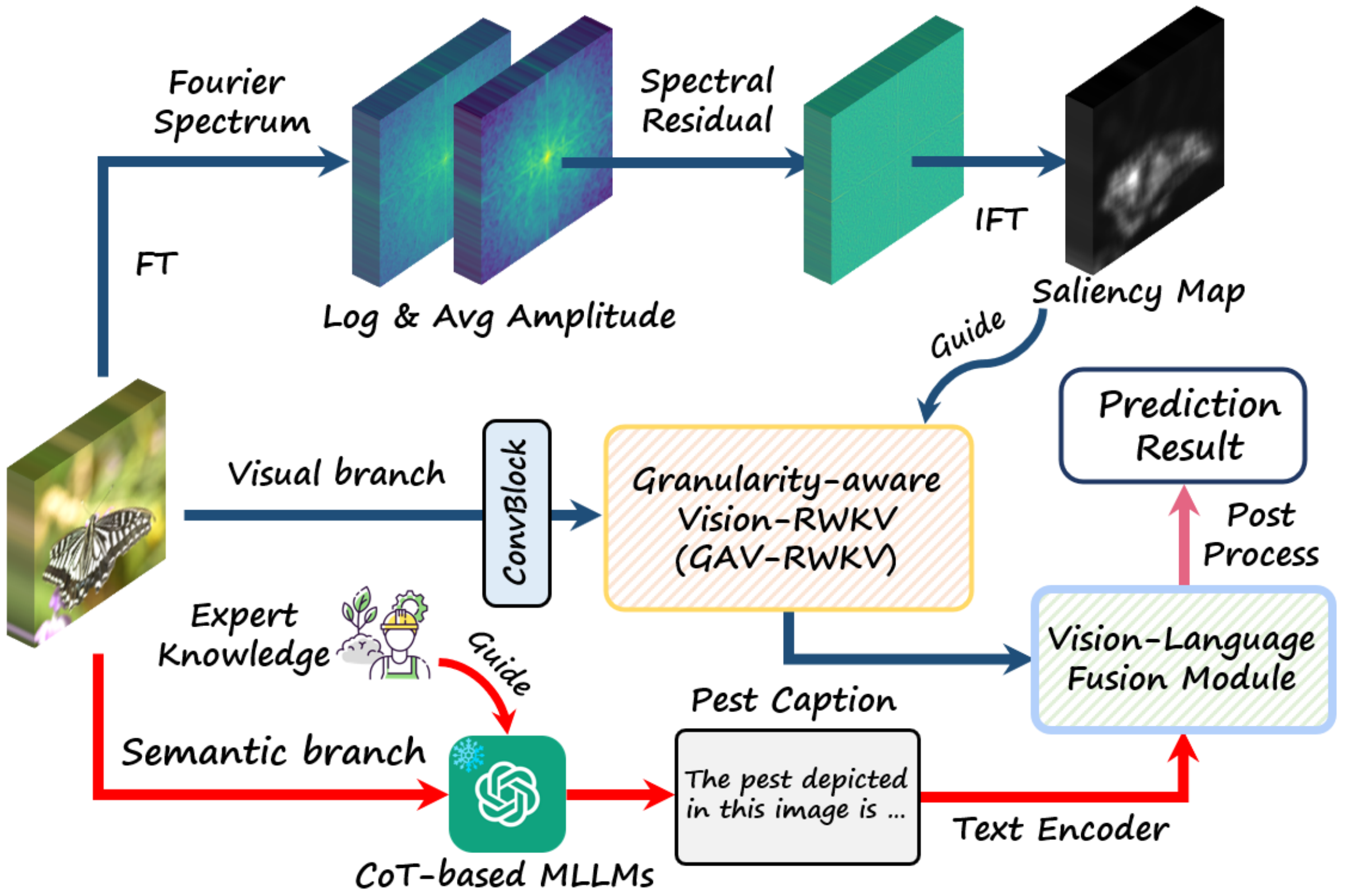}
	\caption{Pipeline for our PestVL-Net. The network comprises two branches: the visual branch leverages Fourier-derived saliency maps to guide granular perception in RWKV, while the semantic branch generates pest-related captions using expert knowledge and priors from MLLMs.}
\label{FIG:1}
\end{figure}

Meanwhile, closely related species often exhibit highly similar visual appearance \cite{han2024crossing}. As illustrated in Fig. \ref{FIG:pest}, such intrinsic visual complexity significantly challenges model learning and impedes accurate recognition, as purely visual analysis struggles to capture the deep semantic features of pests. However, most existing pest image processing approaches prioritize complex visual feature extraction modules \cite{passias2024insect}, neglecting the richer, high‑level semantic information intrinsic to the insect itself. Currently, the  fine-grained  integration of visual and linguistic modalities for multimodal pest analysis remains largely underexplored. 

While multimodal learning typically relies on language annotations for visual datasets, acquiring such annotations manually remains a labor-intensive and costly process. The recent emergence of Multimodal Large Language Models (MLLMs) \cite{bai2025qwen2} has introduced a new paradigm for image semantics and scene understanding. Pre-trained on extensive internet-sourced datasets, these models have demonstrated impressive capabilities in tasks such as image captioning and visual question answering. Consequently, many research \cite{wang2025wcg} has explored leveraging MLLMs for image caption generation and multimodal learning paradigms. However, directly applying MLLMs to generate captions for pest images encounters notable limitations. This is primarily attributed to the inherent difficulty in adapting the generalized knowledge embedded in MLLMs to the precise morphological attributes  specific to pests. Nonetheless, accurate pest description generation remains critical for enabling effective multimodal learning in pest recognition. 

To address the aforementioned limitations and conduct research that contributes to applications in real-world pest management, we photographed and collected two high-definition datasets comprising diverse species of common crop pests for multimodal pest analysis. Building upon above observations, we propose PestVL-Net, a novel multimodal pest learning framework that leverages fine-grained vision-language interaction. For the visual modality, we employ RWKV to extract visual features of pests. To enable the RWKV to effectively capture the core features of pests characterized by rich appearance details, we introduce an improved saliency-guided window partitioning method. Specifically, we employ the spectral residual approach \cite{hou2007saliency} to compute the saliency map of the pest image based on its Fourier amplitude spectrum, as illustrated in Fig. \ref{FIG:1}. Subsequently, we leverage the saliency map to identify and more finely partition information-dense regions, thereby guiding the RWKV to perform detailed fine-grained visual modeling of the pests.  
While for the text modality, we initially leverage the prior knowledge within MLLMs to facilitate the generation of pest image descriptions. To acquire more precise descriptions of pest morphology and texture, we curate and integrate experienced agricultural expert-curated pest knowledge and employ multimodal Chain-of-Thought (CoT) reasoning to guide MLLMs in generating comprehensive pest image captions. We encode the detailed textual information and deeply fuse the derived high-level pest semantic features with the extracted visual features via the cross-attention mechanism, ultimately achieving fine-grained multimodal pest learning.

In summary, our contributions are as follows:
\begin{itemize} 
\item  We propose a novel fine-grained multimodal pest recognition framework based on the adaptive RWKV architecture that integrates the perception of pest morphological information. To the best of our knowledge, this is the first work to introduce RWKV into agricultural research.
\item We introduce two multi-species crop pest datasets and leverage the prior knowledge of MLLMs, enriched with agricultural expert experience and guided by multimodal chain-of-thought reasoning, to generate accurate corresponding pest captions for effective multimodal learning.
\item Extensive experimental evaluations demonstrate the efficacy of our proposed PestVL-Net, achieving superior performance across multiple agricultural pest datasets and highlighting its potential in precision agriculture and pest management.
\end{itemize}

\section{Related Work}
\subsection{DL-based Pest Learning}
In recent years, deep learning approaches have witnessed increasing adoption within the agricultural and pest research communities \cite{qi2025density}. Most existing deep learning–based methods for pest analysis rely on standard backbones and utilize transfer learning to develop task-specific models \cite{odounfa2024deep, hu2024causality}. 
More recently, novel methods have emerged in the agricultural domain that integrate popular network architectures to analyze insects from diverse perspectives. For example, InsectMamba \cite{wang2025insectmamba} introduces Mamba to address insect recognition tasks. Currently, several efforts \cite{nguyen2024insect} have also been made to organize and collect insect datasets for fundamental research. BIOSCAN-5M \cite{gharaee2024bioscan} provides a large-scale biological dataset of insects with multimodal information, including DNA sequences and geographic data, to support tasks like zero-shot transfer and multimodal retrieval. Shinoda et al. \cite{shinoda2025agrobench} introduce AgroBench, a large-scale benchmark for Vision-Language Models in agriculture, which is manually annotated by domain experts to rigorously evaluate model performance across seven real-world scenarios, such as crop, pest, and weed identification and management. However, while such datasets are valuable, they are often oriented toward biological analysis and demand high computational resources.  In contrast, existing crop pest datasets typically cover only a narrow range of species, limiting their utility in real-world pest management scenarios. Currently, there is a scarcity of comprehensive datasets and research for the learning of multi-species common pests.

\subsection{Receptance Weighted Key Value}
The Receptance Weighted Key Value (RWKV) architecture \cite{peng2023rwkv, duan2024vision}, initially conceived for language modeling, has emerged as a computationally efficient paradigm that combines the  capability of parallel processing with the linear scaling advantages. RWKV employs the wkv attention mechanism to model long-range dependencies with linear complexity compared to Transformer-based networks \cite{vaswani2017attention}, thus affording enhanced global contextual awareness. Vision-RWKV \cite{duan2024vision} was the first to introduce RWKV into the visual domain, paving the way for a series of subsequent studies. Since its introduction, RWKV has attracted increasing research attention \cite{li2025freq, yang2025restore, yuan2024mamba}. For instance, RWKV-CLIP \cite{gu2024rwkv} integrates RWKV into visual-language representation learning, while RWKV-SAM \cite{yuan2024mamba} applies it to image segmentation. Furthermore, Restore-RWKV \cite{yang2025restore} pioneered the use of RWKV in medical image restoration  and StyleRWKV \cite{dai2025stylerwkv} has designed an RWKV-based network for style transfer. However, these studies are still in the preliminary stages of exploring the core modeling mechanisms of the RWKV model. In this paper, we explore the RWKV modeling approach for agricultural pest images and propose a saliency-guided, fine-grained window partitioning strategy to enhance RWKV. 
\section{Method}
\begin{figure*}[ht]
	\centering
        \includegraphics[width=\linewidth]{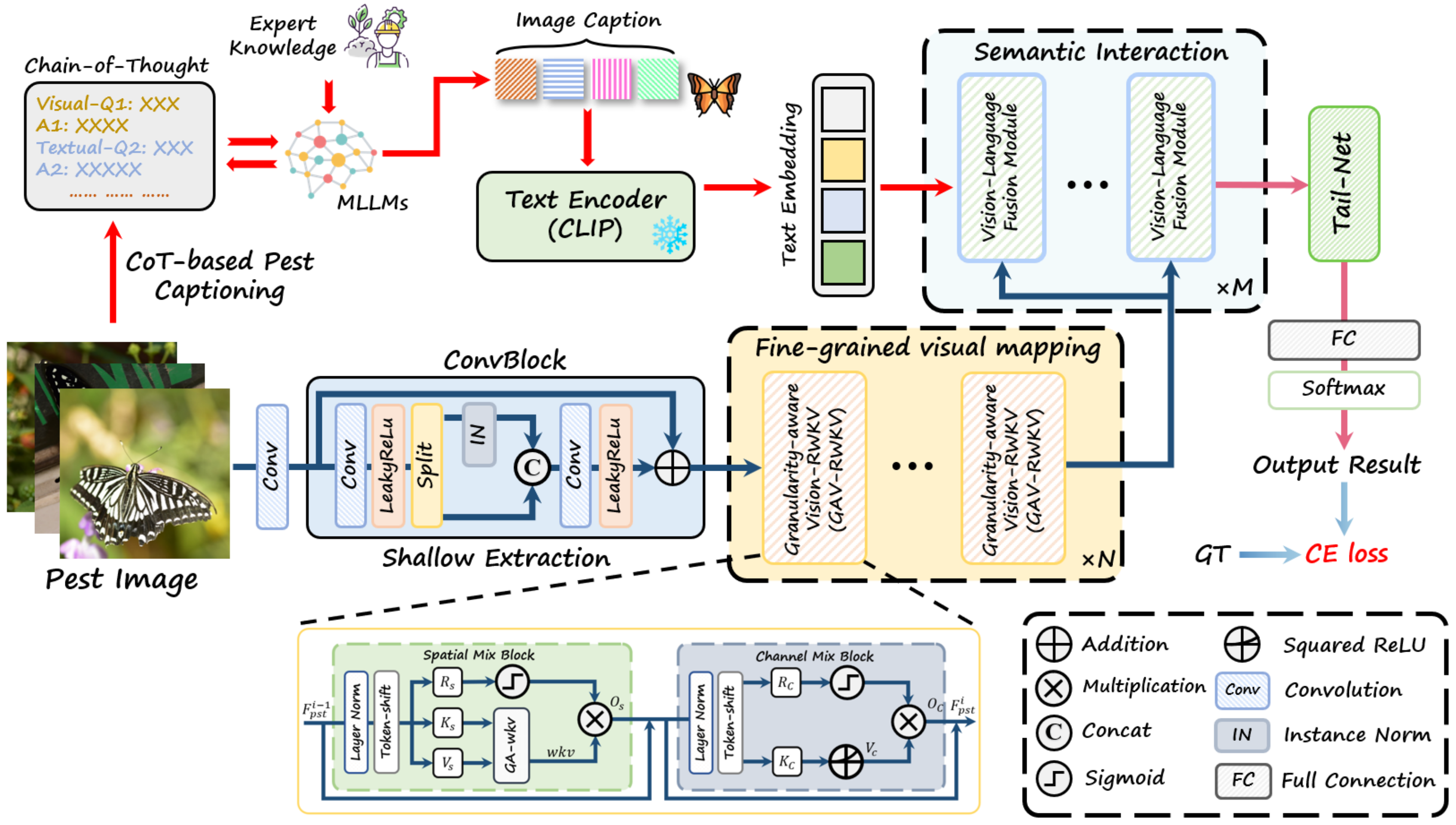}
	\caption{Overall framework of our proposed PestVL-Net, which comprises several main components: fine-grained visual feature modeling, pest semantic generation and encoding, and multimodal feature fusion via semantic interaction.}
\label{FIG:2}
\end{figure*}
\subsection{Spectral Residual} 
The spectral residual approach is primarily used to compute the pest saliency map, as shown in Fig. \ref{FIG:1}. Given a pest image $x\in{R^{h\times w}}$, we first apply the Fourier transform (FT) to obtain its amplitude spectrum. The phase spectrum is preserved solely to enable the inverse Fourier transform (IFT) and is omitted from further discussion. The detailed procedure is as follows:
\begin{align}
&\mathcal{F}(x)(u,v)=\frac{1}{\sqrt{HW}} \sum_{h=0}^{H-1}\sum_{w=0}^{W-1}x(h,w)e^{-j2\pi(\frac{h}{H}u+\frac{w}{W}v)} \\
&\mathcal{A}(x)(u,v)=[R^{2}(x)(u,v)+I^{2}(x)(u,v)]^{\frac{1}{2}}
\end{align}
where $R(x)$ and $I(x)$ represent the real part and imaginary part of the frequency representation of the image, and the $\mathcal{A}(x)$ denote the corresponding amplitude spectrum. Next, we calculate the logarithmic amplitude spectrum $\mathcal{L}(x)$. To estimate redundant background information in the spectrum, we apply a mean filter  to obtain the average log-amplitude spectrum $\mathcal{L}_{avg}(x)$. The spectral residual $\mathcal{R}(x)$ is then obtained by subtracting these two items:
\begin{align}
&\mathcal{L}(x)(u,v)=\ln(A(x)(u,v)+\epsilon), \\
&\mathcal{L}_{avg}(x)(u,v)=\mathcal{L}(x)(u,v)*h_n(x)(u,v), \\
&\mathcal{R}(x)(u,v)=\mathcal{L}(x)(u,v)-\mathcal{L}_{avg}(x)(u,v).
\end{align}
Here, $\epsilon$ is a small constant used to prevent the logarithm from reaching zero and $h_n(x)$ denotes the mean filter with a kernel size of $n \times n$. The spectral residual $R(x)$ is treated as the amplitude spectrum in the frequency domain, which is then combined with the original phase spectrum to perform the IFT, finally resulting in the saliency map $\mathcal{S}_{pst}$.

\subsection{Network Framework}
The overall framework of our proposed PestVL-Net is depicted in Fig. \ref{FIG:2}. At the visual level, the input pest image is passed through the convolutional block \cite{chen2021hinet}  for shallow feature extraction, then subjected to fine-grained mapping in the Granularity-Aware Vision-RWKV (GAV-RWKV) module. Concurrently, at the linguistic level, expert knowledge and Chain-of-Thought (CoT) reasoning are employed to guide the MLLMs in generating a descriptive caption corresponding to the pest image. The visual and linguistic features are deeply integrated within the Vision-Language Fusion (VLF) module and processed by post-processing modules (e.g. tail network) to yield the final predictions.

\subsection{Key Components}
\begin{figure}[ht]
	\centering
        \includegraphics[width=\linewidth]{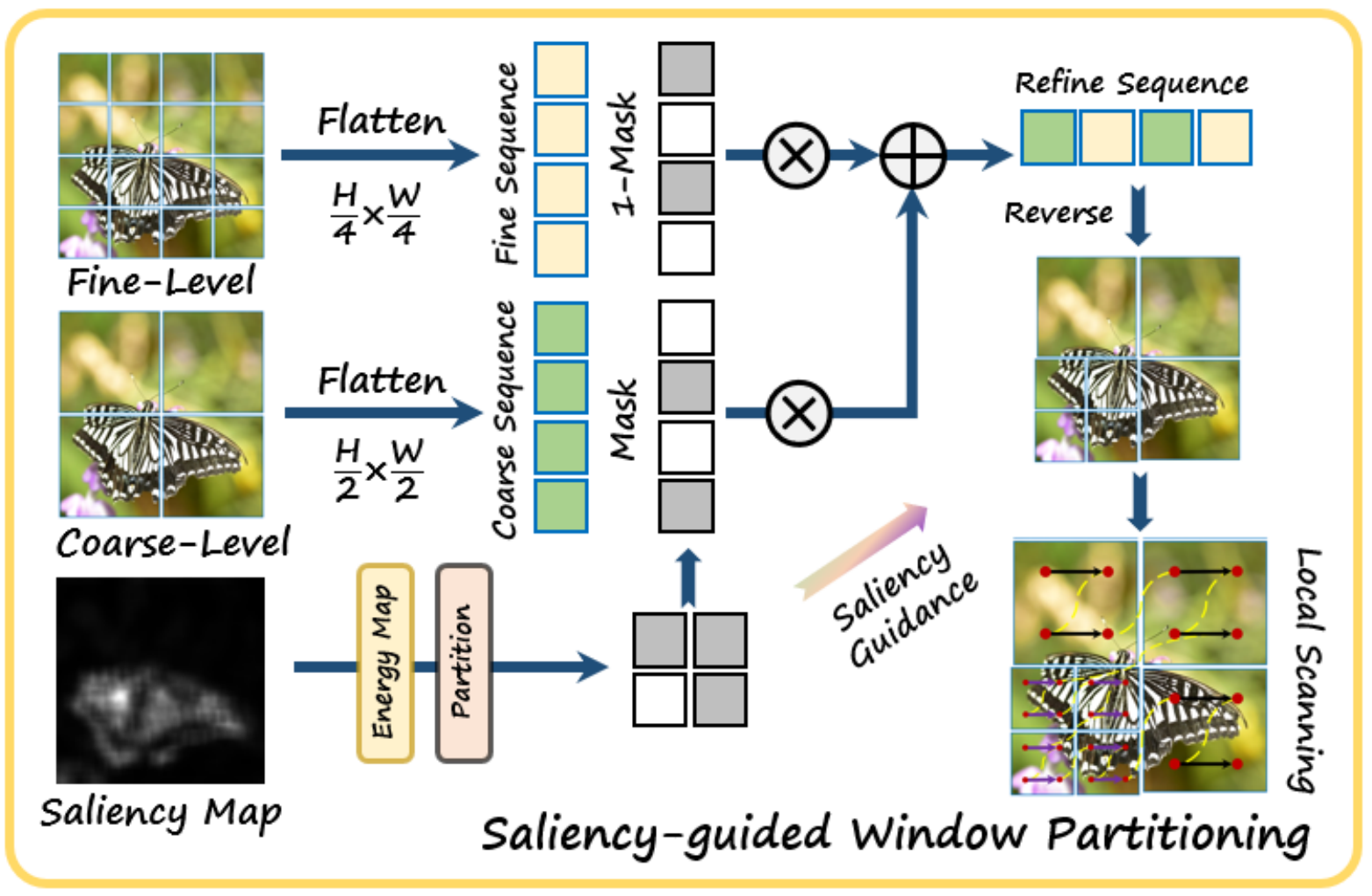}
	\caption{Saliency-guided window partitioning and local scanning within the GAV-RWKV module.}
\label{FIG:3}
\end{figure}

\textbf{Saliency-guided Window Partitioning.} To enable fine-grained visual feature modeling of the complex morphological and textural characteristics of pests, we drawing inspiration from \cite{xie2024quadmamba} and improve the RWKV architecture by introducing the saliency-guided window partitioning mechanism, as illustrated in Fig. \ref{FIG:3}. Specifically, upon obtaining the saliency map via Fourier spectrum, we apply an energy mapping function $\text{EM}(\cdot)$, where regions with higher energy values denote greater saliency. The saliency map enables a more precise and effective modeling of the pests’ key appearance features. We partition the pest image into coarse-level sub-windows $W_{{N}_{1}}$ with ${N}_{1}=4$ and fine-level sub-windows $W_{{N}_{2}}$ with ${N}_{2}=16$. We employ the TopK algorithm to identify the quadrant with the highest energy in the saliency map $\mathcal{S}_{pst}$, which then serves as the basis for subsequent fine-grained window partition:
\begin{align}
& \{s_{{N}_{1}}^{{i}}|i\in[0,N_1)\}=\text{Partition}(\text{EM}(\mathcal{S}_{pst}|W_{{N}_{1}}^{i} )), \\
 & {k~\xleftarrow{index}~\text{TopK}(~s_{N_1}^i~|~K=1),~i\in [0,N_1)}, \\
& \{W_{{N}_{2}}^{({k,j})}|j\in[0,N_1)\}=\text{Partition}(W_{{N}_{1}}^{{k}}).
\end{align}
Here, with $K=1$, we apply coarse-grained partitioning to the saliency energy map and the highest-energy coarse sub-window $W_{{N}_{1}}^{{k}}$ is then further subdivided into fine-grained sub-windows $W_{{N}_{2}}^{({k,j})}$ for more detailed modeling. However, the index-based selection is non-differentiable, impeding end-to-end network training. To address this, we employ the $\text{Gumbel-Softmax}(\cdot)$ to construct a differentiable selection mask over the flattened window sequence:
\begin{align}
& M = \uparrow \{ (\text{Gumbel-Softmax}(\{s_{{N}_{1}}^{{i}}|i \in[0,N_1)\}) \}, \\ 
& L_{N_1} = \text{Flatten}(W_{N_1}), L_{N_2} = \text{Flatten}(W_{N_2}), \\
& L_{pst}=((1-M_{L})\odot L_{N_1})+(M_{L}\odot L_{N_2})
\end{align}
where $L_{N_1}$ and $L_{N_2}$ denote the coarse- and fine-grained token sequences. $M_{L}$ represents the full sequence mask obtained after upsampling and $L_{pst}$ is the resulting fine-grained sequence. We then utilize the inverse window transform to $L_{pst}$ and perform local scanning on each sub-window to reinforce the local dependencies of pest features.  

\noindent \textbf{Granularity-aware Vision-RWKV module.} GAV-RWKV module leverages the aforementioned  saliency–guided window partitioning and scanning process and the detailed architecture is presented in Fig. \ref{FIG:2}. Given input pest feature $F_{pst}^{i-1}$ is first processed by the Spatial Mix Block (SMB) to capture long-range dependencies across spatial dimension. The feature $F_{pst}^{i-1}$ is flattened into sequence $\mathcal{X} \in \mathcal{R}^{T\times C}$, where $T$ denotes the number of tokens and $C$ represents the channels. The sequence $\mathcal{X}$ is then passed through a layer normalization (LN) operation, followed by a shift layer using the depth-wise convolution ($\text{DConv}$): 
\begin{align}
X_{S}=\alpha \text{DConv} \circ \text{LN}(\mathcal{X})+\beta \text{LN}(\mathcal{X})
\end{align}
where $\alpha$ and $\beta$ are the learnable parameters, and the operator $\circ$ denotes continuous inter-layer operation. The shift mechanism employed here mirrors the implementation in \cite{yang2025restore}. The sequence $X_{S}$ is subsequently fed into three parallel linear projection layers, parameterized as $W_{R}, W_{K}$ and $W_{V}$, to obtain receptance ${R_S}$, key ${K_S}$, and value $V_S$: 
\begin{equation}
\begin{aligned}
{R_{S}=X_{S}W_{R}}, 
{K_{S}=X_{S}W_{K}},
{V_{S}=X_{S}W_{V}}
\end{aligned}
\end{equation}
The key ${K_S}$ and value $V_S$  are utilized to compute the WKV attention mechanism. Notably, while standard vision RWKV typically employs a Bi-WKV \cite{duan2024vision} mechanism, our approach  introduces a granularity-aware window partitioning and scanning strategy, leading to a refined mechanism that we denote as Ga-WKV to highlight this distinction. The computation for remaining  sequence tokens follows an analogous procedure:
\begin{align}
wkv_t &= \text{Ga-WKV}(K_S,V_S)_t \nonumber  \\
&=\frac{\sum_{i=1,i\neq t}^Te^{-(|t-i|-1)/T\cdot w+k_i}v_i+e^{u+k_t}v_t}{\sum_{i=1,i\neq t}^Te^{-(|t-i|-1)/T\cdot w+k_i}+e^{u+k_t}}
\end{align}

\noindent where the ${w}$ and ${u}$ are hyperparameters within the attention and $wkv_t$ represents the computed attention output for $t$-th token. The ${k_i}$ and ${v_i}$ denote the $i$-th tokens extracted from key ${K_S}$ and  value $V_S$, respectively. The $wkv$ attention output is multiplied by $R_S$, which is activated by the sigmoid function $\sigma(\cdot)$, and the result is then processed by a linear projection layer $W_{O}$ to generate the SMB ouput $O_{S}$:
\begin{equation}
\begin{aligned}
O_{S}={(\sigma(R_{S})\odot wkv)W_{O}}   
\end{aligned}
\end{equation}
After the residual connection, $O_{S}$ is passed into the Channel Mix Block (CMB) to fuse features along the channel dimension. This block follows a similar structure, comprising a LN layer, a shift operation, and three linear projection layers $W_{R}$, $W_{K}$ and $W_{V}$:
\begin{align}
&{X_{C}=\text{Shift} \circ \text{LN}(O_S) + \mathcal{X}}, \\
&{R_{C}=X_{C}W_{R}}, {K_{C}=X_{C}W_{K}},{V_{C}=\gamma (K_C)W_{V}}.
\end{align}

Notably, the value $V_{C}$ is computed differently from receptance $R_{C}$ and key $K_{C}$, it is obtained by applying a squared ReLU activation $\gamma(\cdot)$ to the $K_{C}$. The final outputs of the CMB and GAV-RWKV module are computed as:
\begin{align}
& O_{C}={(\sigma(R_{C})\odot V_C)W_{O}}, \\
& F_{pst}^{i} = O_{C} + O_{S}.
\end{align}

\noindent \textbf{CoT-based Pest Captioning.} To enhance the network's capacity to learn high-level semantic information regarding  morphological and textural characteristics both within and across species, we introduce a text-driven multimodal learning approach that promotes fine-grained vision–language interaction. We leverage the priors within MLLMs to automatically generate descriptive captions ${T}_{pst}$ for the pest images $x$, as shown in Fig. \ref{FIG:2}, thereby avoiding extensive manual annotation. We  utilize GPT-4o \cite{hurst2024gpt}, known for its robust scene understanding capabilities, as the foundation for our MLLM-based caption generation. Moreover, to augment MLLMs’ comprehension of crop pest appearance features within agricultural contexts, we incorporate the domain-specific knowledge of experienced agricultural experts (${P}_{expert}$) and employ multimodal CoT prompting (${P}_{CoT}$) strategies to guide the generation of these descriptions. Following this generation process, we encode textual description  into the text feature embedding ${F}_t$ using a pre-trained CLIP text encoder, as follows:
\begin{align}
& {T}_{pst}=f_{\text{MLLM}}(x |\{{P}_{expert},{P}_{CoT} \}), \\
& {F}_t = \text{CLIP}({T}_{pst}).
\end{align}
Here, the $f_{\text{MLLM}}(\cdot)$ represents the employed MLLMs. This approach guides the MLLMs to reason through relevant agricultural information, leading to more accurate and contextually rich textual descriptions that can effectively enhance the learning of fine-grained visual representations. The specific visualization example is shown in Fig. \ref{FIG:5}.

%

\noindent \textbf{Vision-Language Fusion module.} 
\begin{figure}[ht]
	\centering
        \includegraphics[width=\linewidth]{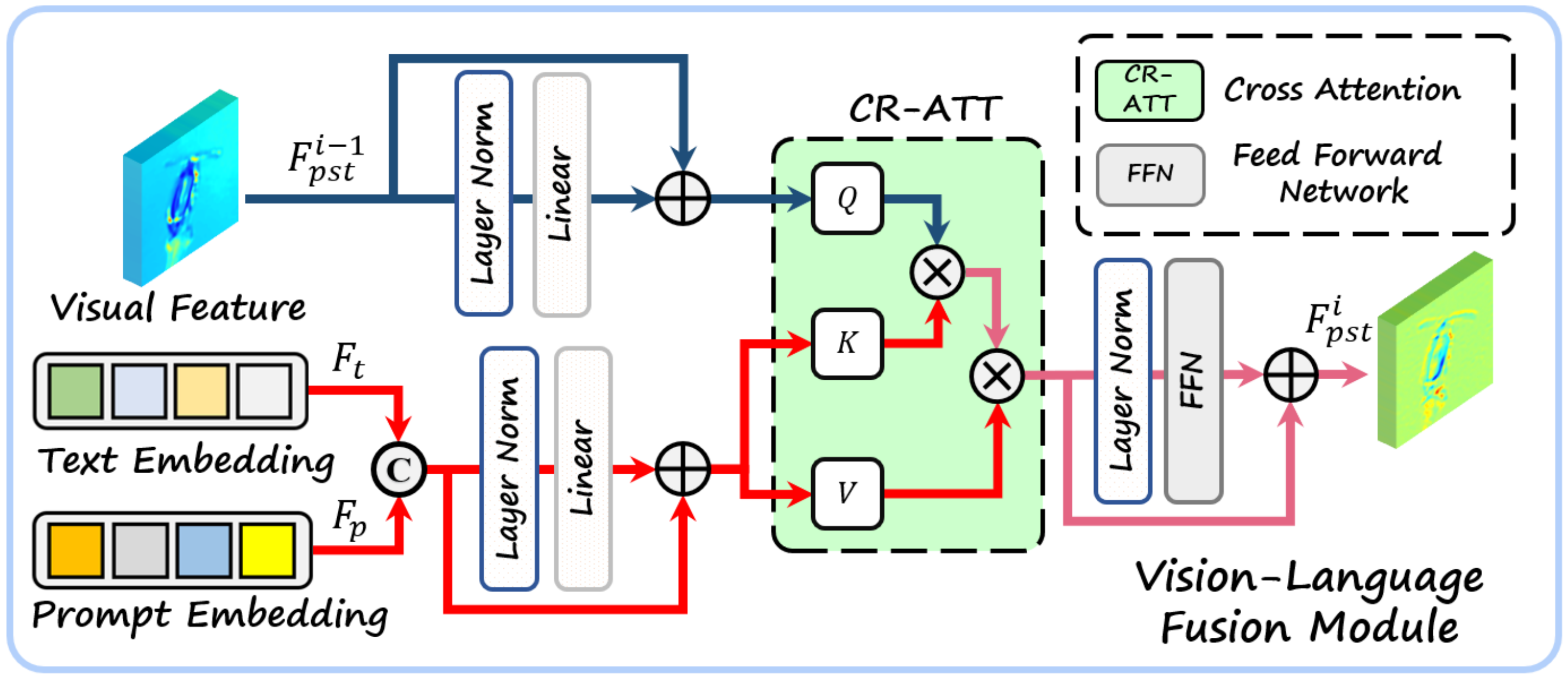}
	\caption{The detailed computational structure of the Vision-Language Fusion module.}
\label{FIG:4}
\end{figure}
Within the VLF module, the pest visual features and the encoded linguistic features undergo a process of deep semantic interaction. Given the input pest visual features $F_{pst}^{i-1}$ and the text embedding ${F}_t$, we further incorporate a learnable prompt embedding $F_p$, which is concatenated with ${F}_t$ to adaptively refine the semantic representation. These features are first independently processed through LN and linear layers ${W}_{linear}$, combined with residual connections for calculation:
\begin{align}
& {\tilde{F}}_{pst}={W}_{linear}\circ \text{LN}({F}_{pst}^{i-1})+{F}_{pst}^{i-1},\\
& F_{text} = [{F}_{t}, {F}_{p}], \\
& {\tilde{F}}_{t}={W}_{linear}\circ \text{LN}(F_{text}) + F_{text}.
\end{align}
Subsequently, the visual feature ${\tilde{F}}$ and textual feature ${\tilde{F}}$ are fused via the cross attention mechanism. The detailed process is outlined as:
\begin{align}
&{Q}_{v} ={\tilde{F}}_{pst}{W}_{qv}, \\
&{K}_{t}, {V}_{t} ={\tilde{F}}_{t}{W}_{kt},{\tilde{F}}_{t}{W}_{vt}, \\
&\widehat{F}_{pst} = \text{CR}({Q}_{v},{K}_{t},{V}_{t}), \\
&\text{CR}(\cdot)=\text{softmax}\left(\frac{{Q}_v {K}_t}{d_k}\right){V}_{t}
\end{align}
where ${W}_{qv}, {W}_{kt}$ and ${W}_{vt}$ denote the corresponding linear projection layers, and $d_k$ is the scaling factor. ${Q}_{v}$ represents the visual query, while ${K}_{t}$ and ${V}_{t}$ stand for the textual key and value, respectively. The resulting fused feature $\widehat{F}_{pst}$ is passed through the feedforward network $\text{FFN}(\cdot)$ to produce the final output of the VLF module:
\begin{align}
{F}_{pst}^{i}=\text{FFN} \circ \text{LN}(\widehat{F}_{pst})+\widehat{F}_{pst}.
\end{align}
The deeply integrated multimodal features resulting from the VLF module are then passed through post-processing components (e.g. tail network, fully connected layers), to yield the final model predictions. With these components integrated, we establish the complete  framework.

\section{Experiments}

\subsection{Datasets}
In this paper, we employ three pest datasets to validate the effectiveness of our proposed method. Specifically, the Li dataset \cite{li2020crop} comprises ten pest species, encompassing a total of 5865 images. The dataset is partitioned into training, validation, and test sets with a ratio of 7:1:2, respectively. QianFSD and AgriInsect are high-definition pest datasets that we photographed and collected, containing 143 and 200 common crop pest species, respectively. These two collections exhibit a richer diversity of pest species, with QianFSD consisting of 7,054 images and AgriInsect containing 9,452 images. Their data division follows a similar ratio to that of the Li dataset. As illustrated in Table \ref{tab:dataset}, we compare with other widely used datasets such as Farm Insects \cite{FarmInsect}, Agricultural Pests  \cite{Agri_Insect}, Insect Recognition  \cite{xie2015automatic}, Forestry Pest \cite{liu2022dataset} and the IP102 \cite{wu2019ip102}. Our datasets provide broader coverage of common pest species and offer stronger support for practical agricultural research.
\subsection{Baseline Methods}
To evaluate the effectiveness of our proposed method, and in light of the limited prior research in this domain, we conduct comprehensive comparisons with a diverse range of basic and advanced deep learning models. Specifically, the benchmark methods include eleven DL-based methods: AlexNet \cite{krizhevsky2014one}, VGG16 \cite{simonyan2014very}, GoogLeNet \cite{szegedy2016rethinking}, ResNet-18 \cite{he2016deep}, ResNet-50 \cite{he2016deep}, ResNeXt-50 \cite{xie2017aggregated}, Swin-Transformer \cite{liu2021swin}, Uniformer \cite{li2023uniformer, li2022uniformer}, SeaFormer \cite{wan2025seaformer++, wan2023seaformer}, DFL \cite{hu2024causality}, VMamba \cite{liu2024vmamba} and TransXNet \cite{lou2025transxnet}.

\begin{table}[ht]
\caption{Comparative statistics of eight pest datasets. The high-quality pest benchmark we constructed advances beyond conventional pest recognition, enabling a more diverse and fine-grained understanding of pest characteristics.}
\centering
\begin{tblr}{
  column{even} = {c},
  column{3} = {c},
  column{5} = {c},
  hline{1,10} = {-}{0.08em},
  hline{2,8} = {-}{0.05em},
}
Dataset                             & Category & Train  & Val  & Test   \\
Farm Insects                       & 15       & 1070   & 153  & 305    \\
Agricultural Pests                 & 12       & 3846   & 549  & 1099   \\
Insect Recognition & 24       & 966    & 138  & 276    \\
Forestry Pest & 31      & 5014    & 716  & 1433   \\
Li dataset                          & 10       & 4106   & 563  & 960    \\
IP102                             & 102      & 45095 & 7508 & 22619 \\
\textbf{QianFSD (Ours)}                & 143      & 4938   & 705  & 1411   \\
\textbf{AgriInsect~\textbf{(Ours)}} & 200      & 6616   & 945  & 1891   
\end{tblr}

\label{tab:dataset}
\end{table}
\begin{table*}[ht]
\caption{Quantitative comparison on three crop pest datasets. Best results highlighted in bold. Suboptimal results are highlighted in \underline{underline}. Each evaluation metric is expressed as a percentage (\%), and higher values indicate better performance.}
\centering
\resizebox{\textwidth}{!}{
\begin{tblr}{
  cells = {c},
  cell{1}{1} = {r=2}{},
  cell{1}{2} = {c=4}{},
  cell{1}{6} = {c=4}{},
  cell{1}{10} = {c=4}{},
  vline{2-3,7} = {1}{0.05em},
  vline{6,10} = {2}{0.05em},
  vline{2,6,10} = {1-15}{0.05em},
  hline{1} = {1}{0.03em},
  hline{1} = {2-14}{0.08em},
  hline{2} = {2-14}{0.03em},
  hline{3,15} = {-}{0.05em},
  hline{16} = {-}{0.08em},
}
Methods       & Lidataset      &                &                &                & QianFSD          &                &                &                & AgriInsect     &                &                &                \\
              & Accuracy       & Precision      & F1 Score       & GM             & Accuracy       & Precision      & F1 Score       & GM             & Accuracy       & Precision      & F1 Score       & GM             \\
AlexNet       & 60.80          & 59.50          & 59.42          & 60.57          & 54.87          & 55.95          & 54.87          & 46.64          & 64.14          & 65.16          & 64.33          & 62.31          \\
VGG16         & 48.37          & 45.61          & 46.08          & 44.74          & 43.46          & 43.01          & 42.93          & 28.90          & 43.72          & 44.13          & 43.86          & 33.14          \\
GoogLeNet           & 54.89              & 53.87              & 54.26              & 54.44              & 48.76              & 50.21              & 49.15              & 46.91              & 52.19              & 53.72              & 52.74              & 52.92             \\
ResNet-18     & 77.98          & 78.73          & 77.91          & 77.33          & 69.03          & 70.73          & 69.44          & 68.50          & 80.29          & 81.95          & 80.90          & 82.00          \\
ResNet-50     & 71.48          & 71.74          & 71.15          & 69.58          & 68.09          & 69.74          & 68.68          & 67.12          & 80.34          & 81.68          & 80.74          & 82.91          \\
ResNeXt-50    & 71.48          & 72.11          & 71.45          & 70.51          & 67.22          & 68.93          & 67.86          & 65.71          & 79.20          & 80.35          & 79.54          & 81.95          \\
Swin-T        & 75.23          & 74.67          & 74.73          & 75.73          & 68.02          & 70.71          & 69.05          & 68.64          & 82.24          & 83.28          & 82.48          & 84.93          \\
UniFormer     & 71.56              & 70.77              & 70.72              & 71.63              & 76.50             & 78.10              & 76.93              & 76.56              & 65.75              & 67.53             & 65.98             & 59.01              \\
SeaFormer     & 77.81              & 77.99              & 77.52              & 77.22              & 72.90              & 74.13              & 73.11              & 73.92              & 83.66              & 84.71              & 83.95              & 85.08              \\
DFL           & \underline{81.65}          & \underline{81.94}          & \underline{81.28}          & \underline{82.49}          & 76.30          & 78.74          & 77.14          & 78.68          & \underline{85.13}          & \underline{86.80}          & \underline{85.74}          & \underline{86.70}          \\
VMamba-B           & 63.64             & 63.33              & 63.20              & 62.10             & 65.62              & 67.80             & 66.37              & 62.82             &  66.75           & 69.24              & 67.72            & 66.87             \\
TransXNet     & 76.90              & 76.31              & 76.27              & 76.59              & \underline{79.44}              &  \underline{80.36}            & \underline{79.56}               & \underline{81.86}             & 80.67              & 81.81              & 81.04              & 81.95              \\
\textbf{Ours} & \textbf{88.49} & \textbf{89.36} & \textbf{88.52} & \textbf{88.56} & \textbf{86.72} & \textbf{87.53} & \textbf{86.86} & \textbf{87.14} & \textbf{90.15} & \textbf{90.88} & \textbf{90.19} & \textbf{90.51} 
\end{tblr}
}
\label{tab:1}
\end{table*}
\subsection{Implement Details}
We trained our model using the PyTorch framework on a single NVIDIA A100 GPU. Generally, the optimization was carried out using the SGD optimizer with an initial learning rate of $0.1$. We trained the model for 200 epochs on the Li dataset and 400 epochs on both the QianFSD and AgriInsect datasets. Notably, each baseline method is trained from scratch using input images resized to 224 × 224, ensuring a fair and consistent comparison. We employed four widely adopted metrics, including Accuracy, Precision, F1-score \cite{goutte2005probabilistic}, and Geometric Mean (GM) \cite{wu2019ip102}, to  evaluate the overall performance. In our proposed PestVL-Net, the network module parameters are set to $N=5$ and $M=2$.

\subsection{Comparison with State-of-the-Art Methods}
The evaluation results of our approach on three pest datasets are shown in Table \ref{tab:1}. Experimental results demonstrate that our proposed network not only significantly outperforms classical deep learning architectures but also exceeds state-of-the-art methods, yielding substantial improvements across all evaluation metrics. On the Lidataset, our method achieves an improvement of about 7\% across multiple metrics, with a particularly notable gain in Precision over the suboptimal method. Consistently strong performance is also observed on the multi-species pest datasets QianFSD and AgriInsect, underscoring the robustness of our approach across diverse pest scenarios.

\subsection{Ablation Experiments}
To investigate the contribution of various components within PestVL-Net, we conduct the ablation experiments on the Lidataset, as summarized in Table \ref{tab:2}. Additional ablation studies conducted on the QianFSD and AgriInsect datasets are provided in the supplementary material.

\noindent \textbf{Convolution and RWKV Architecture.} To assess the contribution of RWKV's superior global receptive field to visual feature extraction, we replaced the RWKV module with a purely convolutional counterpart, as specified in config (I) of Table \ref{tab:2}. The results clearly indicate a significant performance decline with the convolutional architecture, underscoring the limitations of traditional convolution module in visual feature modeling. This also highlights the effectiveness of RWKV in capturing long-range dependencies, contributing to more accurate pest recognition.

\begin{table}[ht]
\centering
\caption{Ablation study results on the Lidataset. }
\begin{tblr}{
  cells = {c},
  cell{1}{1} = {r=2}{},
  cell{1}{2} = {c=4}{},
  vline{2} = {1,2,3-8}{0.05em},
  hline{1,9} = {-}{0.05em},
  hline{2} = {2-5}{0.03em},
  hline{3,8} = {-}{0.05em},
}
\textbf{Config} & \textbf{Lidataset} &        &        &        \\
       & Accuracy         & Precision   & F1 Score    & GM   \\
(I)    & 81.65       & 82.07 & 81.19 & 82.30 \\
(II)    & 85.15            & 85.79     & 85.12       & 85.45     \\
(III)    & 86.24            & 86.47     & 85.93    & 85.58      \\
(IV)    & 84.65            & 85.36      & 84.69      & 84.71      \\
(V)    & 87.91            & 88.52      & 87.81      & 87.40     \\
Ours   & \textbf{88.49} & \textbf{89.36} & \textbf{88.52} & \textbf{88.56} 
\end{tblr}

\label{tab:2}
\end{table}
\noindent \textbf{Saliency-guided Window Partitioning.} In config (II), we conducted an ablation study on the saliency-guided window partitioning and scanning strategy within the GAV-RWKV module. The experimental results demonstrated a performance degradation of approximately 3\% across various metrics, affirming the efficacy of this mechanism for fine-grained modeling of pest visual features.

\begin{figure*}[ht]
	\centering
        \includegraphics[width=\linewidth]{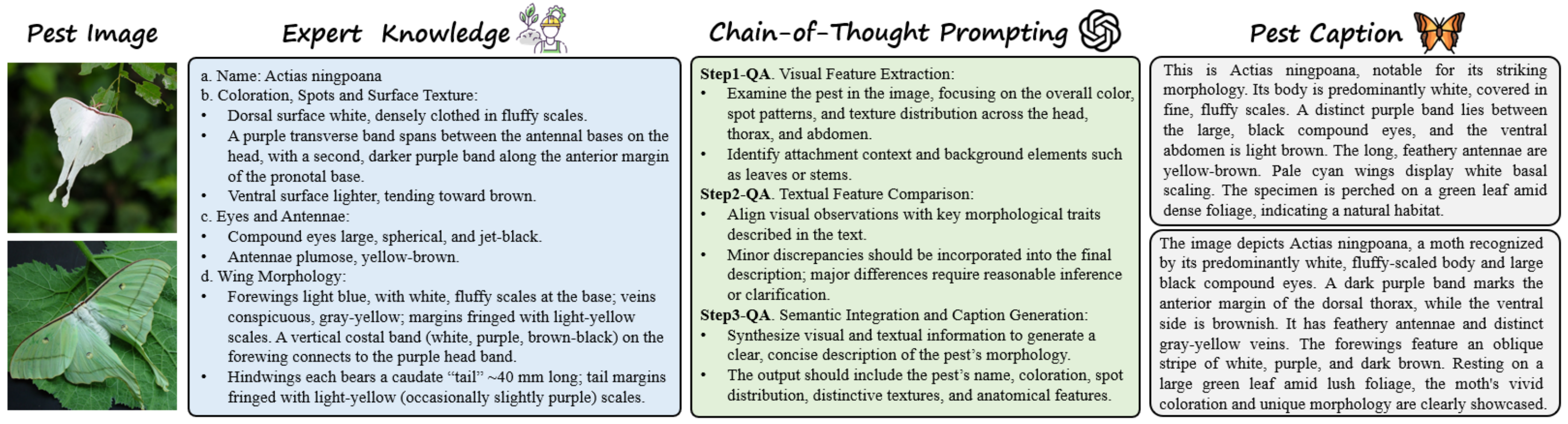}
	\caption{The progress for generating pest captions leveraging agricultural expert knowledge and multimodal Chain-of-Thought reasoning, illustrated with examples from the AgriInsect dataset. The same approach is applied to the other two datasets.}
\label{FIG:5}
\end{figure*}

\noindent \textbf{Chain-of-Thought Guidance.} To investigate the influence of expert knowledge and Chain-of-Thought guidance on pest semantic generation, we  employed MLLMs for identifying and generating captions for pest images, specifically in a setting where such auxiliary information was omitted. As illustrated in config (III), the resulting degradation in model performance demonstrates that general MLLMs struggle to effectively capture the intrinsic morphological and contextual characteristics of pests in agricultural scenes without domain-specific reasoning guidance.

\noindent \textbf{Semantic Information Fusion.}
As part of the ablation study, we removed the Vision-Language Fusion module to  evaluate the contribution of semantic information. The notable performance degradation observed in config (IV) of Table \ref{tab:2} demonstrates the positive impact of incorporating semantic information through the proposed multimodal fusion strategy centered on pest feature descriptions.

\noindent \textbf{Learnable Prompt Embedding.} In Config (V), we examined the effect of learnable prompt embedding $F_p$ capacity. The prompt embeddings adaptively learn semantic characteristics and serve a corrective function in the presence of semantic bias.  The slight performance drop observed upon the removal of $F_p$ highlights its contribution to alleviating biased semantic representations.

\subsection{Visualization}
\noindent \textbf{Captioning Methodology.} Fig. \ref{FIG:5} illustrates an example of pest caption generation on the AgriInsect dataset, guided by expert agricultural knowledge and multimodal CoT reasoning.  Starting with an input pest image, we first compile relevant expert knowledge, including key visual attributes like color, spots, and texture. This structured information is subsequently utilized as CoT-based prompts to guide the MLLMs in generating semantically rich and contextually accurate captions. The same knowledge-driven methodology is applied across all three multimodal pest datasets to ensure coherent and domain-aligned caption generation.

\begin{figure}[t]
	\centering
        \includegraphics[width=\linewidth]{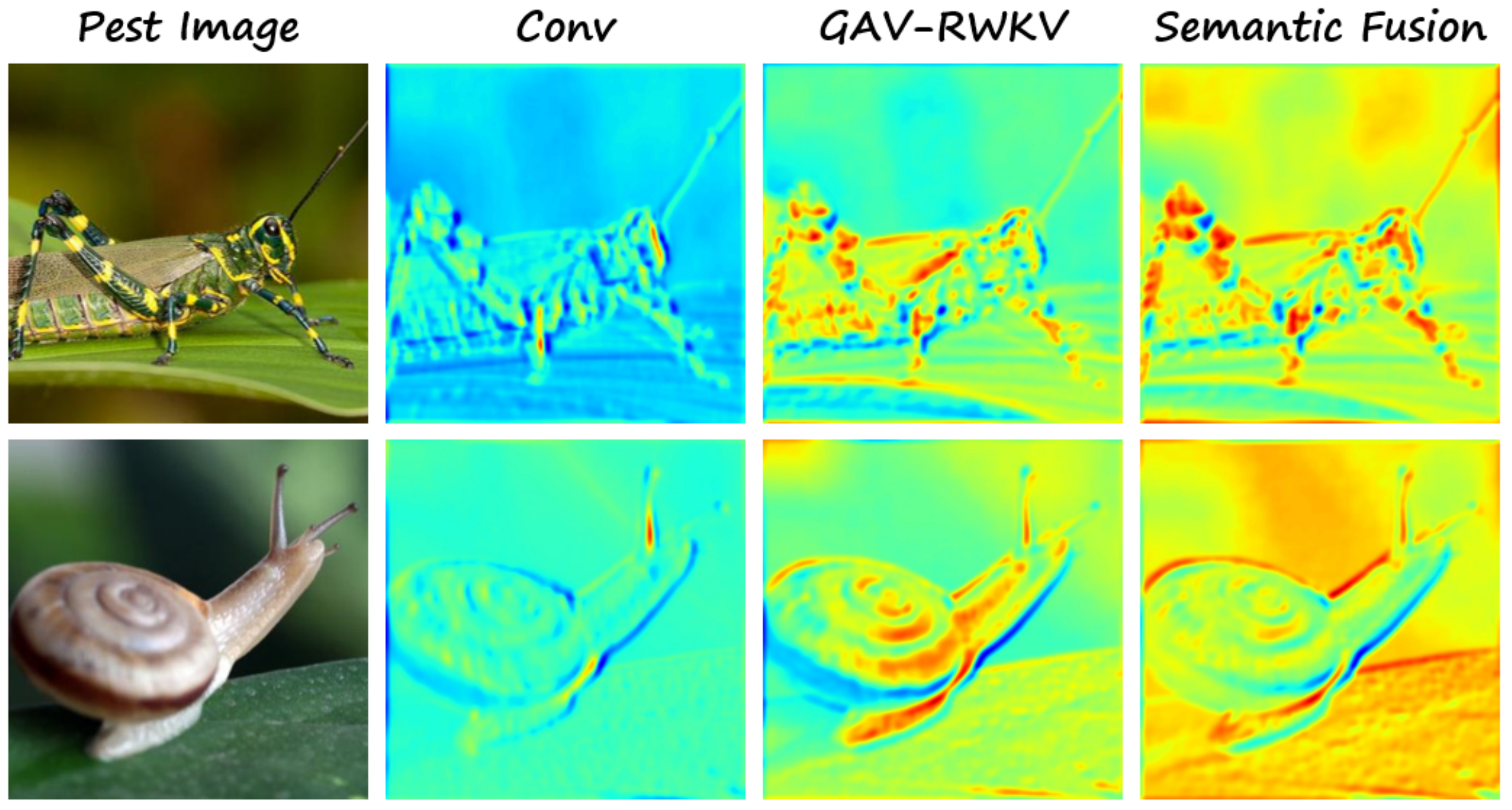}
	\caption{Visualization of feature maps generated by distinct modules on the Lidataset.  The intensity of warm colors (e.g., yellow and red) directly correlates with the degree of model attention allocated to specific regions.}
\label{FIG:7}
\end{figure}

\noindent \textbf{Feature Map Analysis.} As shown in Fig. \ref{FIG:7}, feature map visualizations validate our module designs. While purely convolutional blocks struggle to capture pest morphology, the GAV-RWKV module successfully maps fine-grained visual features. This attention is further refined after semantic fusion, proving the efficacy of our approach.

\section{Conclusion}
In this paper, we address the challenges posed by the intricate morphological and textural characteristics of pests in agricultural scenes, further compounded by the inherent visual heterogeneity across pest species, which significantly hinder effective pest representation learning. To this end, we propose a saliency-guided visual RWKV module that enables adaptive image window partitioning and hierarchical feature extraction. The resulting visual representations are seamlessly integrated with high-level pest semantics derived from MLLMs leveraging the Chain-of-Thought reasoning and agricultural expert knowledge, thereby facilitating fine-grained and context-aware multimodal pest understanding. Extensive experiments conducted on diverse pest datasets further demonstrate the effectiveness and robustness of the proposed framework.

\section*{Acknowledgements}
This work was supported by the Natural Science Foundation of Anhui Province, China (No. 2208085MC57) and Central-Oriented Foundation for Local Science and Technology Development, China (No. 202407a12020010).

{
    \small
    \bibliographystyle{ieeenat_fullname}
    \bibliography{main}
}


\end{document}